\documentclass[letterpaper, 10 pt, conference]{ieeeconf}
\IEEEoverridecommandlockouts

\usepackage{amsmath,amsfonts}
\usepackage{pifont}

\usepackage{bm}
\usepackage{algorithmic}
\usepackage{algorithm}

\usepackage[caption=false]{subfig}

\usepackage{array}
\usepackage{textcomp}
\usepackage{stfloats}
\usepackage{url}
\usepackage{verbatim}
\usepackage{graphicx}

\usepackage{booktabs}
\usepackage{tabularx}
\usepackage{makecell}  

\usepackage{cite}

\makeatletter
\let\NAT@parse\undefined
\makeatother

\usepackage[colorlinks,linkcolor=black,anchorcolor=black,citecolor=black,urlcolor=black,hyperfootnotes=true]{hyperref}   
\usepackage[all]{hypcap} 

\begin{document}

\title{\LARGE \bf
Potato: A Data-Oriented Programming 3D Simulator \\for Large-Scale Heterogeneous Swarm Robotics
}

\author{Jinjie Li$^{1}$, Liang Han$^{2*}$, Haoyang Yu$^{2}$, Zhaotian Wang$^{2}$, Pengzhi Yang$^{3}$, Ziwei Yan$^{2}$, Zhang Ren$^{1}$
\thanks{$^{1}$J. Li and Z. Ren are with the School of Automation Science and Electrical Engineering, Beihang University, Beijing, 100191, China
{\tt\small \{lijinjie, renzhang\}@buaa.edu.cn}}
\thanks{$^{2}$L. Han, H. Yu, Z. Wang, and Z. Yan are with the Sino-French Engineer School, Beihang University, Beijing, 100191, China
{\tt\small \{liang\_han, haoyang\_yu, wangzhaotian, yanziwei \}@buaa.edu.cn}}
\thanks{$^{3}$P. Yang is with the Faculty of Electrical Engineering, Mathematics and Computer Science, Delft University of Technology, Delft, 2628 CD, Netherlands
{\tt\small P.Yang-4@student.tudelft.nl}}
}

\maketitle

\begin{abstract}
Large-scale simulation with realistic nonlinear dynamic models is crucial for algorithms development for swarm robotics. However, existing platforms are mainly developed based on Object-Oriented Programming (OOP) and either use simple kinematic models to pursue a large number of simulating nodes or implement realistic dynamic models with limited simulating nodes. In this paper, we develop a simulator based on Data-Oriented Programming (DOP) that utilizes GPU parallel computing to achieve large-scale swarm robotic simulations. Specifically, we use a multi-process approach to simulate heterogeneous agents and leverage PyTorch with GPU to simulate homogeneous agents with a large number.
We test our approach using a nonlinear quadrotor model and demonstrate that this DOP approach can maintain almost the same computational speed when quadrotors are less than 5,000. We also provide two examples to present the functionality of the platform.
\end{abstract}


\section{Introduction}


Swarm robot systems can accomplish tasks that individual robots cannot complete alone through cooperation and coordination, which has recently received extensive attention from academia and industry. Developing perception, planning, and control algorithms for these swarm robots requires experiments on hardware systems to verify their effectiveness. However, field experiments for swarm robots are demanding to conduct due to challenges such as large experimental sites, high maintenance difficulty, and high failure rate. Therefore, utilizing a simulator with realistic models is necessary to verify the swarm algorithms.

The simulators suitable for swarm robots have requirements in two dimensions: the number of simulation nodes and the fidelity of the simulation models. However, existing robotic simulators usually focus on one dimension. Some robotic simulators attempt to achieve large-scale simulations at the expense of fidelity, such as the simple kinetic model in BeeGround \cite{lim_beeground_2021} and the modified unicycle model in SCRIMMAGE \cite{demarco_simulating_2019}.
Other popular robotic simulators provide realistic models while supporting the simulation of only about 50 robots on a desktop computer, such as AirSim \cite{shah_airsim_2017} and Gazebo-based RotorS \cite{furrer_rotorsmodular_2016}. The above simulators leverage mainly central processing units (CPUs) for numerical computation, which limits their capability for large-scale simulation.

\setlength{\dbltextfloatsep}{8pt plus 1.0pt minus 2.0pt}
\begin{figure}[t]
    \centering
    \subfloat[Object-Oriented Programming (OOP)]{
        \includegraphics[trim=0 5 0 0,clip,width=0.9\linewidth]{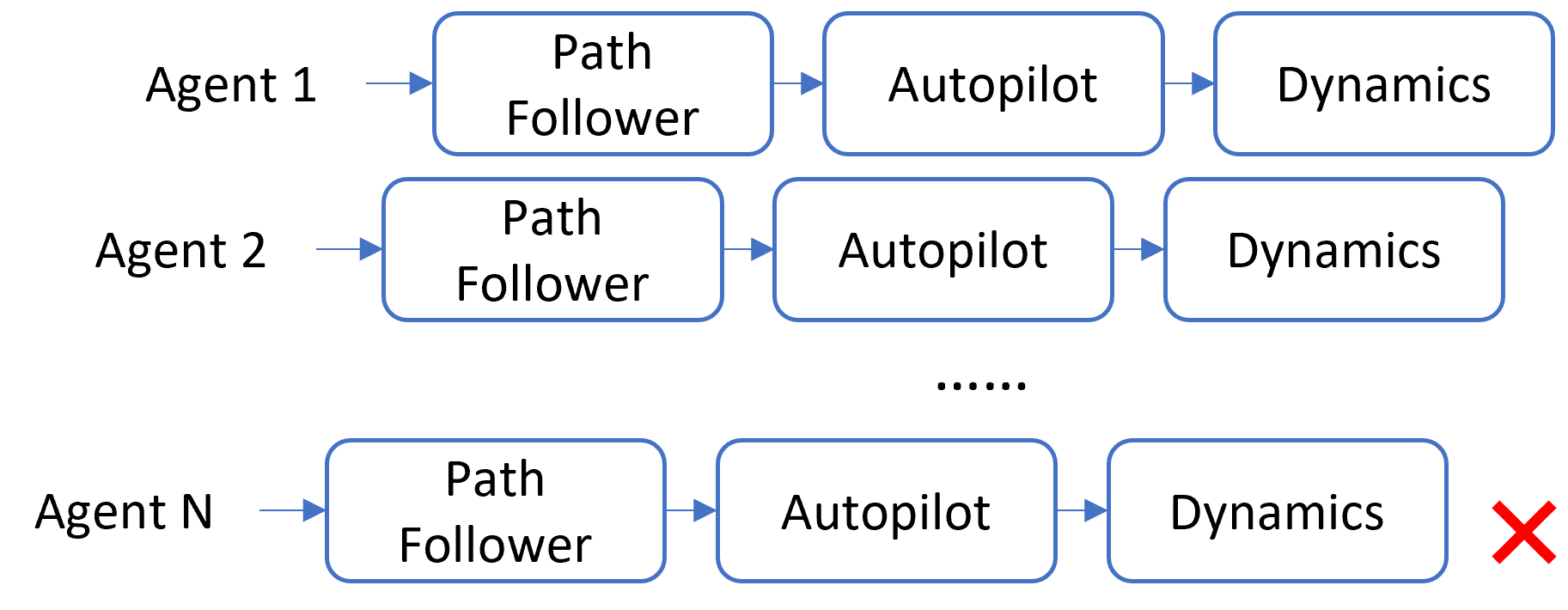}
        \label{fig:oop}
    } \\
    \subfloat[Data-Oriented Programming (DOP)]{
        \includegraphics[trim=0 10 30 25,width=0.9\linewidth]{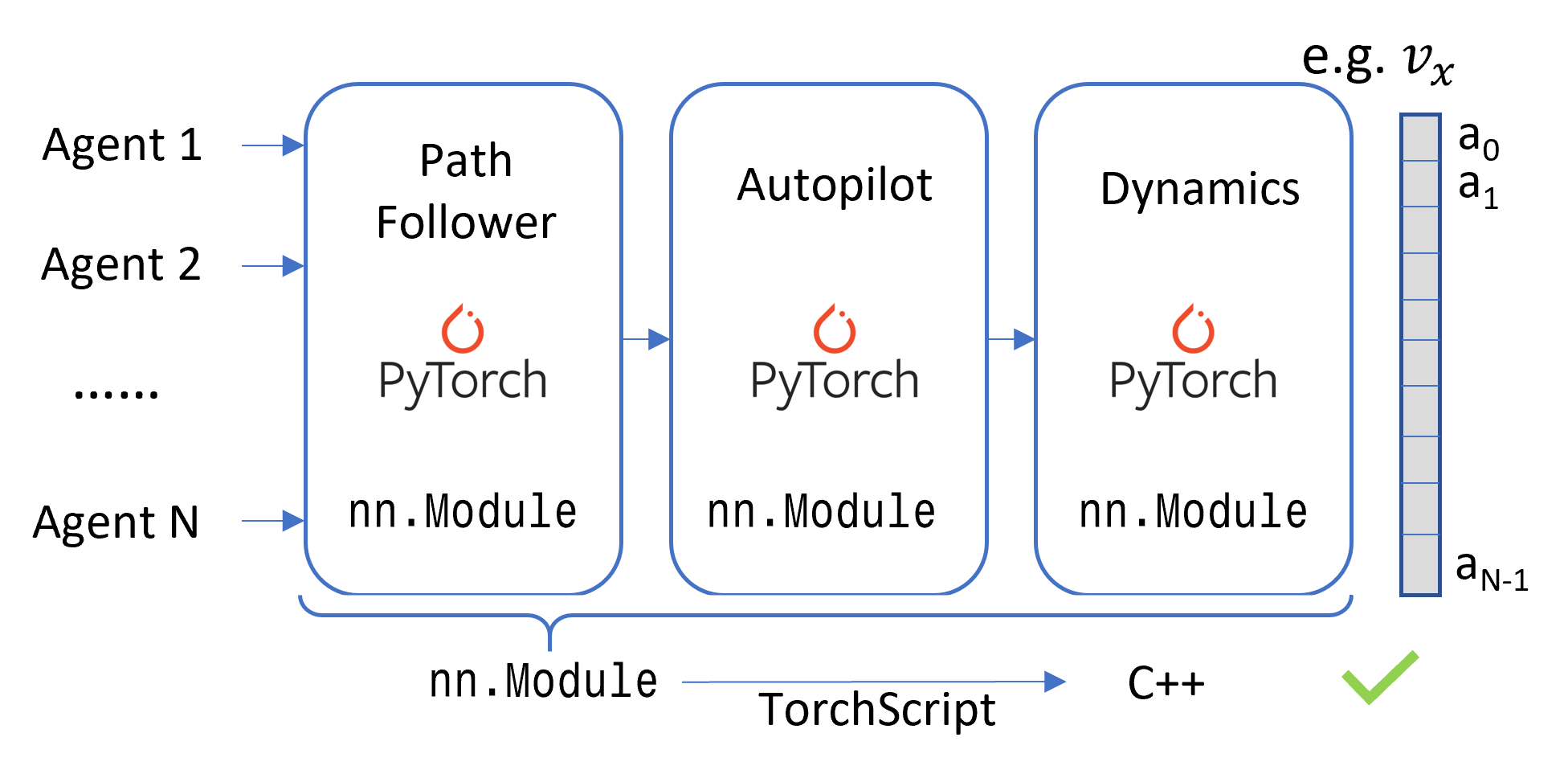}
        \label{fig:dop}
    }
    \caption{The core idea of our simulator. Traditional robotic simulators are developed using OOP, where multiple agents are multiple instances as shown in (a). The agents are then computed through for-loops, multi-threads, or multi-processes, etc. However, since a desktop CPU typically has 10-20 threads, 
    each CPU thread computes multiple agents serially in a loop for large-scale simulation. As a result, the computational speed increases almost linearly with the number of agents. In contrast, the proposed simulator is developed using DOP, grouping the computations of homogeneous agents together and parallelizing them in batches using tensors as shown in (b), which can be computed directly on GPUs. In addition, each computation module can be written as an \texttt{nn.Module} and compiled into C++ using PyTorch's TorchScript for further acceleration. This approach maintains almost the same computational speed for the number of agents below a certain level (below 5,000 in our test).}
    \label{fig:oop_dop}
\end{figure}

Alternatively, the advancement of graphics processing units (GPUs) has opened up the potential for conducting large-scale and high-quality simulations in parallel. NVIDIA's Isaac Gym \cite{makoviychuk_isaac_2021} is an example simulation tool that utilizes GPUs to parallelly simulate the physical world, indicating that GPUs can handle nonlinear models with high fidelity for large-scale simulations.
However, Isaac Gym primarily aims at the algorithm development of deep reinforcement learning, which provides an interface differing from the requirement of swarm robotics. In addition, Isaac Gym is developed using CUDA and C++, which is difficult to master and modify internally, making it challenging for scientific research on swarm algorithms. These limitations are considered when developing the proposed simulator.

In this paper, we develop \textbf{Potato}, a GPU-based large-scale heterogeneous robot simulator. Unlike traditional OOP-based simulators, Potato is designed using DOP, naturally supporting the numerical computation of large-scale nodes on GPUs.
Compared with Isaac Gym, our platform is developed using Python and PyTorch and has cross-platform compatibility as well as lower development difficulty. Furthermore, the simulation can be accelerated using TorchScript, a tool in PyTorch used for deep learning acceleration. Finally, we conduct experiments to verify the effectiveness of the proposed architecture, and we present two demos based on this platform. We hope the idea proposed in this paper can promote the development of the next-generation swarm robotic simulator.



\section{Methodology}

\setlength{\textfloatsep}{8pt plus 1.0pt minus 2.0pt}  
\begin{figure*}[t]
    \centerline{\includegraphics[trim=0 0 0 0,clip,width=\textwidth]{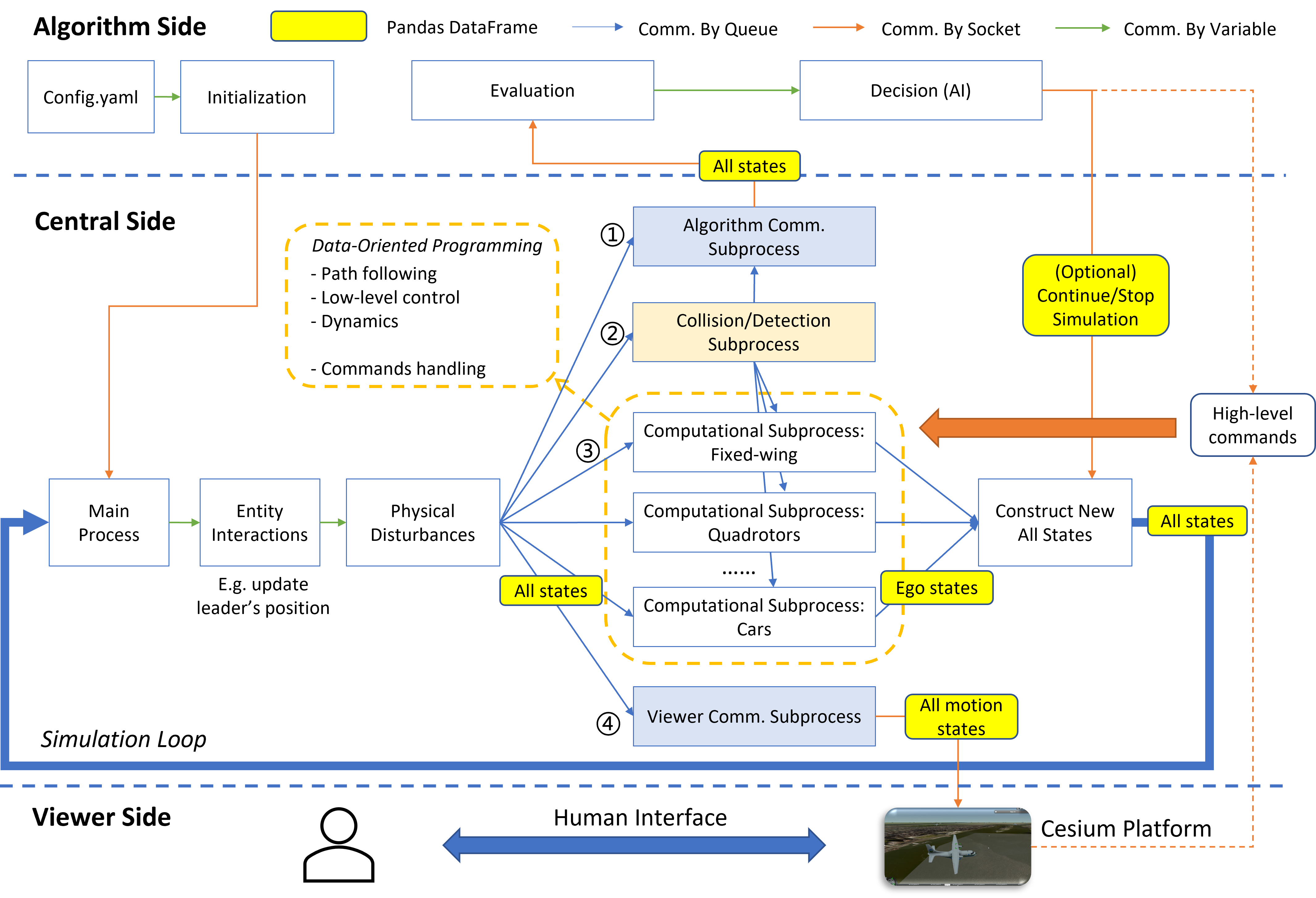}} 

    \caption{The system structure of the proposed simulator. The proposed simulator consists of a Simulation Loop where the states of all agents are transmitted to four directions. Direction \ding{172} sends the states to the \textit{algorithm side} via an algorithm communication subprocess, which uses this information for evaluation and decision making. The generated commands are then sent to computational subprocesses for handling. Direction \ding{173} calculates collision and detection results, which are also sent to computational subprocesses for handling. Direction \ding{174} computes low-level algorithms and dynamics for heterogeneous agents, and sends the updated states back to the main process to refresh the all-states data. Finally, Direction \ding{175} uses a viewer communication subprocess to visualize all agents' motions, and users can manipulate the mouse to influence the agents' behaviors.}
    \label{fig:system}
\end{figure*}

\subsection{System Architecture}
This section introduces the system architecture of the developed large-scale heterogeneous simulation platform, as shown in Fig. \ref{fig:system}. From top to bottom, the entire system is divided into three sides: an \textit{algorithm side}, a \textit{central side}, and a \textit{viewer side}. The \textit{algorithm side} mainly generates decision-making instructions according to the states of the agents; the \textit{central side} controls the simulation process and computes the ordinary differential equations (ODEs) of dynamics; the \textit{viewer side} displays the movement of agents and servers as a human-computer interface. Socket communication is used between each end, making these ends capable of running on different computers. Furthermore, the simulator written in Python can run on different operating systems.


The whole simulation process is depicted in Fig. \ref{fig:system}. At the beginning, the \textit{algorithm side} sets the simulation parameters by reading a configuration file. Then, during each simulation loop, all agents' states (stored in a \texttt{pandas.DataFrame}) are circulated in different system modules, and each module changes all or part of the agents' states.

Unlike other simulation platforms, we take the collision/detection module and \textit{algorithm side} out of the main simulation loop based on the following considerations: First, keeping as few modules as possible in the main loop accelerates the computational speed. Second, taking the decision module (\textit{algorithm side}) outside of the loop approximates the real world. In the real world, humans make decisions with the changing physical world, so the physical world can run for a while before receiving decision-making instructions. 
Third, taking the collision/detection module outside can still guarantee the correct result as long as its computational speed is similar to the main loop. Then the collision information is handled in the main loop using an event-triggered mechanism, and colliding agents will be marked as dead and excluded from the entire system.
We also retain the option of putting these modules into the main loop.



\subsection{Quadrotor Dynamics \& Control}

\setlength{\textfloatsep}{8pt plus 1.0pt minus 2.0pt}
\begin{figure}[t]
    \centerline{\includegraphics[trim=0 0 0 0,clip,width=3.0in]{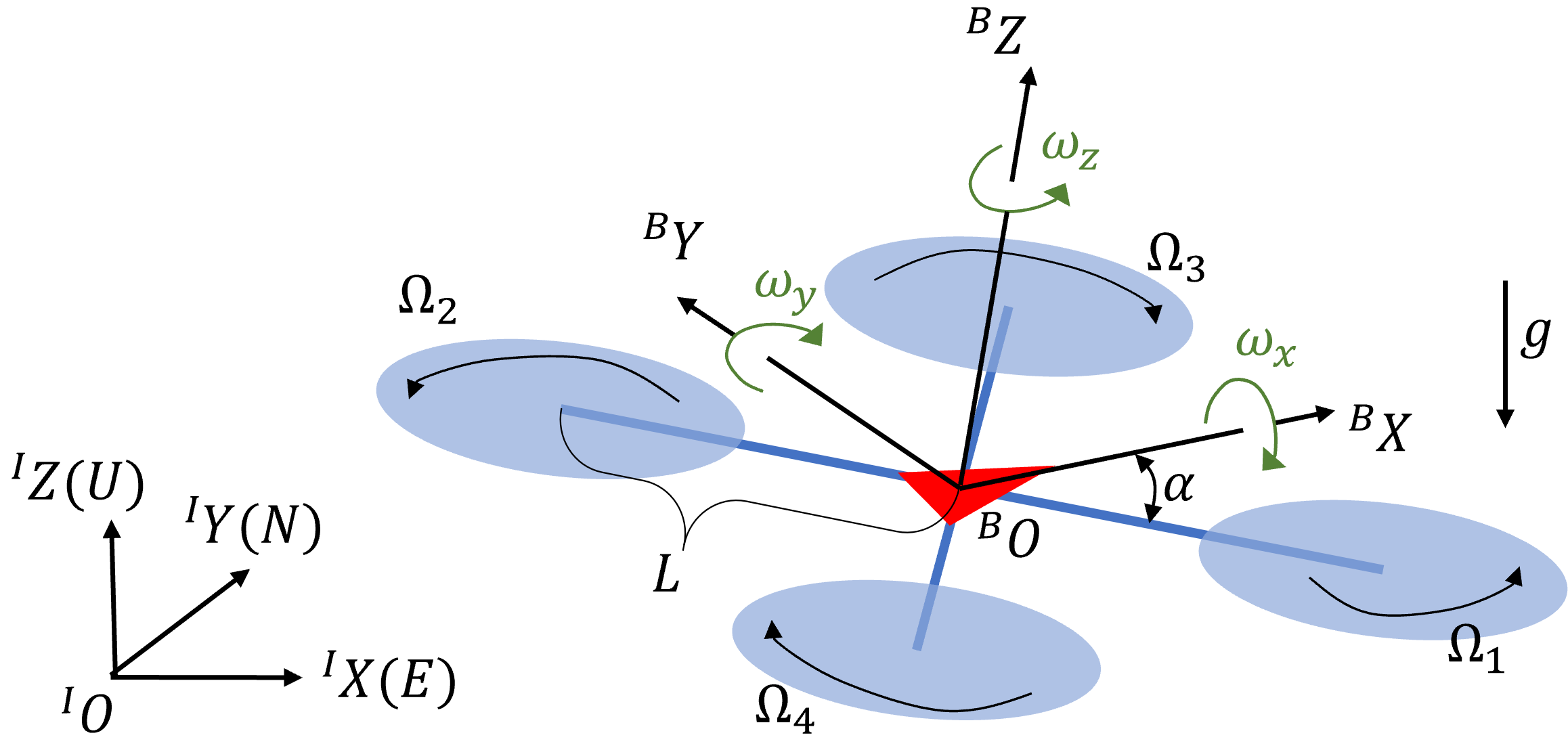}} 
    \vspace*{-1mm}
    \caption{Diagram of the quadrotor model with the ENU (X East, Y North, Z Up) inertial frame and the FLU (X Forward, Y Left, Z Up) body frame.}
    \label{fig:coordinate}
\end{figure}

Three types of mobile robots have been implemented in this simulator, including fixed-wing drones \cite{beard_small_2012}, quadrotors \cite{sun_comparative_2022}, and cars \cite{kabzan_learning-based_2019}. The quadrotors are utilized to test the performance and hence are briefly introduced here.

We assume that the origin of the body frame $\mathcal{B}$ is at the center of mass, and four rotors are all placed in the $\mathcal{B}$ frame's XY-plane. Established from 6-DoF rigid-body dynamics, the quadrotor model is written as follows
\begin{align}
{^I\dot{\boldsymbol{p}}} &= {^I\boldsymbol{v}}, \label{eq:qd1} \\[5pt]
{^I\dot{\boldsymbol{v}}} &= \left({^{I}_{B}\boldsymbol{R}(\boldsymbol{q})} \cdot {^B\boldsymbol{f}_u} \right)/ {m} + {^I\boldsymbol{{g}}}, \label{eq:qd2}\\[5pt]
{^I_B\dot{\boldsymbol{q}}} &= 1/2 \cdot {^I_B\boldsymbol{q}} \circ \left[ \begin{array}{c}
0 \\ 
^B\boldsymbol{\omega} 
\end{array} \right], \label{eq:qd3}\\[5pt] 
{^B\dot{\boldsymbol{\omega}}} &=\boldsymbol{{I}}^{-1} \cdot\left(-{^B \boldsymbol{\omega}} \times\left({\boldsymbol{{I}}} \cdot {^B \boldsymbol{\omega}}\right)+{^B\boldsymbol{\tau}_u} \right), \label{eq:qd4}
\end{align}
where $\circ$ indicates quaternion multiplication, $m$ is mass, ${^I\boldsymbol{{g}}}=[0,0,{-g}]^T$ is  gravity vector, $\boldsymbol{{I}}=\texttt{diag} ({I}_{xx}, {I}_{yy}, {I}_{zz})$ is inertia matrix assuming that the quadrotor exhibits symmetry across all three axes, ${^B\boldsymbol{f}_u}$ and ${^B\boldsymbol{\tau}_u}$ are force and torque caused by the rotors, and ${^B\boldsymbol{\omega}}=\left[\omega_x,\omega_y,\omega_z \right]^T$ is angular rate vector expressed in the body frame.

The thrust generated by rotors is assumed to be vertical to the $\mathcal{B}$ frame's XY-plane, and we therefore obtain ${^B\boldsymbol{f}_u}=\left[0,0,f_c \right]^T$ and ${^B\boldsymbol{\tau}_u}=\left[\tau_x, \tau_y, \tau_z \right]^T$, where $f_c$ is the collective force of four rotors. We use a quadratic fit to model the thrust and torque for each propeller:
\begin{equation}
    f_i={k}_{t} \cdot \Omega^2, \quad \tau_i={k}_{q} \cdot \Omega^2,
    \label{eq:2}
\end{equation}
where ${k}_{t}$ and ${k}_{q}$ are the thrust coefficient and torque coefficient, respectively, as well as $\Omega$ represents motor speed in RPM.
Then the $\left[f_c, \tau_x, \tau_y, \tau_z \right]^T$ and the thrust of each rotor $f_i$ is connected by
\begin{equation}
    \left[f_c, \tau_x, \tau_y, \tau_z \right]^T = \boldsymbol{G} \cdot \left[f_1, f_2, f_3, f_4 \right]^T,
    \label{eq:3}
\end{equation}
in which the control allocation matrix $\boldsymbol{G}$ is
\begin{equation}
\boldsymbol{G}=\left[\begin{array}{cccc}
1 & 1 & 1 & 1 \\
{L} \sin \mathbf{\alpha} & -{L} \sin \mathbf{\alpha} & -{L} \sin \mathbf{\alpha} & {L} \sin \mathbf{\alpha} \\
-{L} \cos \mathbf{\alpha} & -{L} \cos \mathbf{\alpha} & {L} \cos \mathbf{\alpha} & {L} \cos \mathbf{\alpha} \\
{k}_{q} / {k}_{t} & -{k}_{q} / {k}_{t} & {k}_{q} / {k}_{t} & -{k}_{q} / {k}_{t}
\end{array}\right],
\end{equation}
where ${L}$ and ${\alpha}$ are geometric parameters shown in Fig. \ref{fig:coordinate}. Finally, the model is discretized by the 4-order \textit{Runge-Kutta} method for numerical simulation.

We also implement a PID body rate controller as the inner control loop. This dynamics \& control model is utilized to test the computational speed in the next section, and we recommend \cite{sun_comparative_2022} to interested readers for more details.




\section{Performance}

\begin{table}[t]
\centering
\caption{Running Time Mean and Standard Deviation (SD) per Round}
\begin{tabular}{ c|c|c|c } 
 \toprule
  \textbf{\makecell{Language \\ Version}} & \textbf{\makecell{PyTorch \\ Version}} & \textbf{\makecell{FALLBACK \\ Error}} & \textbf{\makecell{Running Time [ms] \\ (mean $\pm$ SD)}} \\
 \midrule
 origin Py3.8 & 1.10.0+cu102 & N & $\textbf{0.8452} \pm 0.0196$ \\ 
 conda Py3.8 & 1.10.0+cu102 & N & $0.8550 \pm 0.0127$ \\
 conda Py3.9 & 1.13.0+cu116 & Y & $1.6974 \pm 0.0141$ \\
 conda Py3.9 & 2.0.0+cu117 & N & $1.4171 \pm 0.0187$ \\
 conda Py3.10 & 1.12.0+cu116 & Y & $1.1738 \pm 0.0248$ \\
 conda Py3.9 & 1.12.0+cu116 & Y & $1.1857 \pm 0.0071$ \\
 conda Py3.8 & 1.12.0+cu116 & N & $0.8678 \pm 0.0081$ \\
 C++ Release & 1.12.0+cu116 & Y (much) & $2.8181 \pm 0.0460$ \\
 C++ Debug & 1.12.0+cu116 & Y (much) & $2.8017 \pm 0.0318$ \\
 \bottomrule
\end{tabular}
\label{tab:run_time_version}
\end{table}

In this section, we test the computational performance of the DOP structure on simulating swarm quadrotors. The performance is tested using a desktop computer with an Intel i7-10700 CPU and an NVIDIA GTX 1660 SUPER GPU. The test program runs on the Ubuntu 20.04 operating system. 

The proposed method relies on TorchScript provided by PyTorch to accelerate the computational speed, and the whole simulation loop can be implemented by Python or C++, so we test the running time for each round under different languages and PyTorch versions. In each test, we fix the number of quadrotors to 1,000 and first run 500 rounds for stable running, then run 2,000 rounds and calculate the average consuming time. Finally, we execute three times for each test and list the result in Table \ref{tab:run_time_version}.

From the table, the C++ version is not faster than the Python version, and we infer the possible reason is many FALLBACK warnings when loading the TorchScript model using \texttt{libtorch}, the C++ API of PyTorch.

\setlength{\textfloatsep}{8pt plus 1.0pt minus 2.0pt}
\begin{figure}[t]
    \centerline{\includegraphics[trim=0 0 0 0,clip,width=0.9\linewidth]{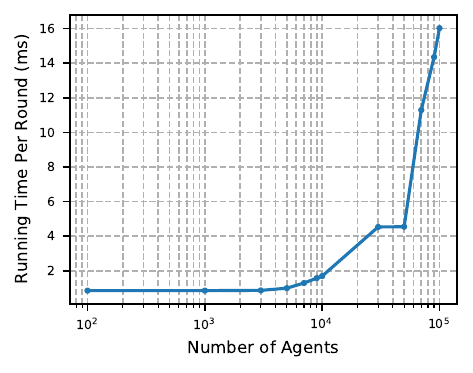}} 
    \caption{Running time as a function of number of agents based on the log coordinates. The time remains stable under 5,000 quadrotors.}
    \label{fig:run_time}
\end{figure}

Then we choose the conda Python 3.8 environment with PyTorch 1.12.0+cu116 to test the change of running time concerning the number of quadrotors, as shown in Fig. \ref{fig:run_time}. The figure shows that the computational time stays almost the same under 5,000 agents and remains less than 2ms for even 10,000 agents, demonstrating the advantage of the DOP method in simulating large-scale agents.

\section{Examples and Extensions}

This section presents two demos of our simulator.

\begin{figure}[htbp]
    \centering
    \subfloat[Demo 1: 1,000 airplanes are flying along circle trajectories, and the trajectories are represented as colorful circles.]{
        \includegraphics[trim=0 0 0 0,clip,width=\linewidth]{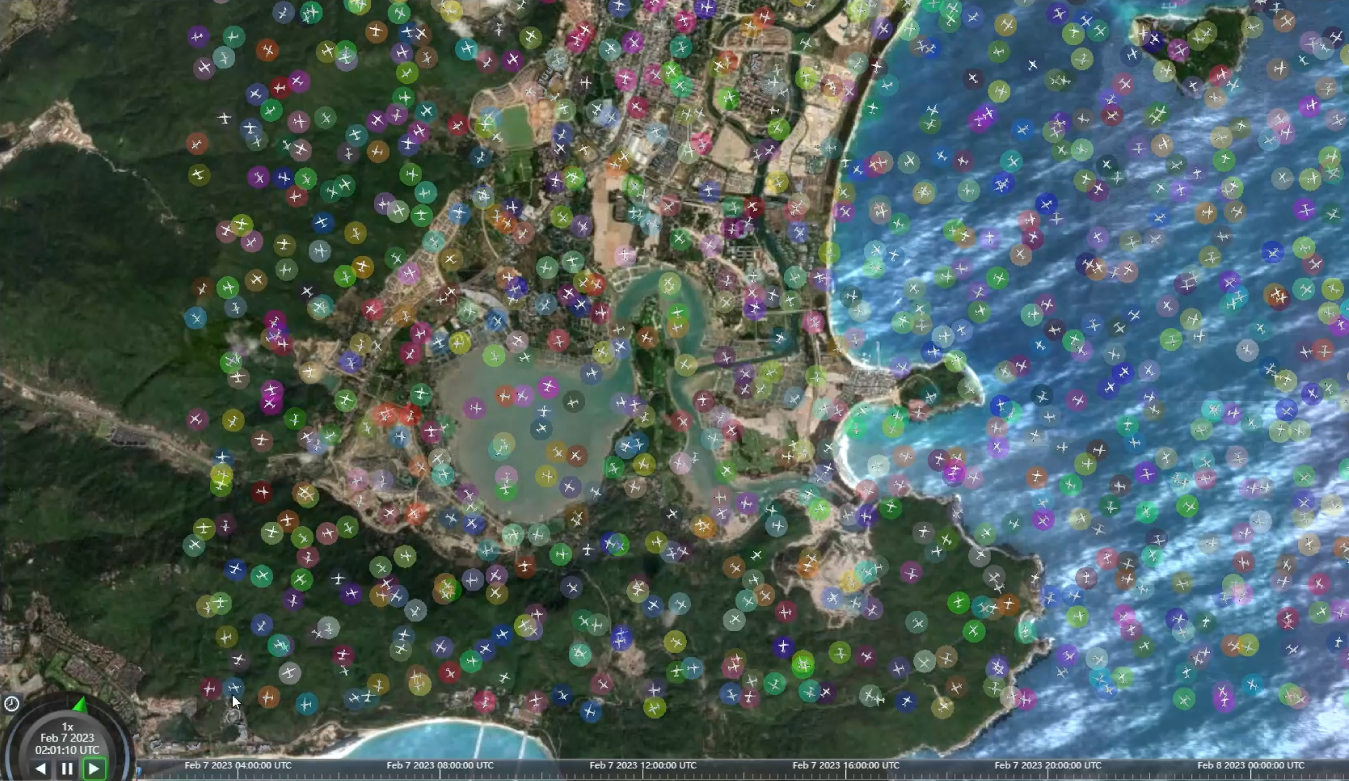}
        \label{fig:demo_1}
    } \\
    \subfloat[Demo 2 \cite{yan_event-triggered_2023}: four real quadrotors and hundreds of virtual quadrotors are flying to verify the formation algorithm in a mixed-reality experimental platform, of which the virtual part is provided by the proposed simulator.]{
        \includegraphics[width=\linewidth]{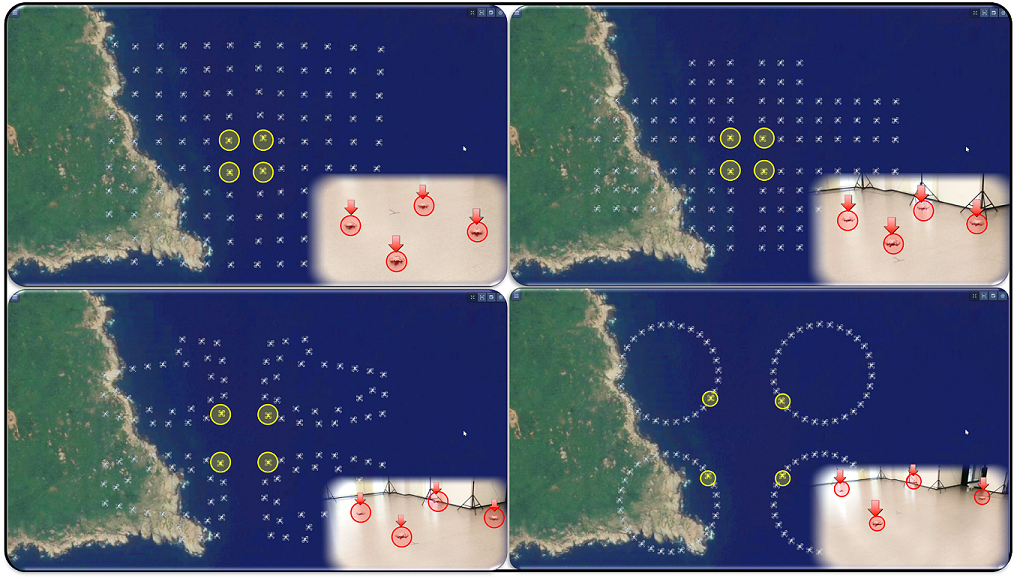}
        \label{fig:demo_2}
    }
    \caption{Two demos to present the functions of our simulator.}
    \label{fig:demos}
\end{figure}

The first demo (Fig. \ref{fig:demo_1}) verifies that our simulator can support over 1000 homogeneous agents simulating on one desktop computer with a GPU. Furthermore, we demonstrate that the simulator can support over 1000 heterogeneous agents, even though the simulating rate is 0.8x.

The second demo shows one research paper using the proposed simulator to verify the algorithms for a large-scale swarm system. Generally, swarm algorithms are demanding to test on large-scale real robots due to limitations in the experimental field and equipment. Thus, this paper builds a mixed-reality platform to verify the effectiveness of the algorithm by applying it equally to limited real robots and hundreds of virtual robots. These virtual robots are supported by our platform, and the positions of both real robots and virtual robots are presented together in our platform, as shown in Fig. \ref{fig:demo_2}.

We plan to apply this platform to more research and present more examples in the future.

\setlength{\dbltextfloatsep}{8pt plus 1.0pt minus 2.0pt}

\section{Conclusion}

In this paper, we developed Potato, a large-scale swarm robotic simulator based on the DOP approach. This simulator used a multi-process approach to simulate different types of agents, and also utilized DOP to accelerate the computation of large-scale homogeneous agents in each process. We leveraged the PyTorch library and TorchScript tool from the deep learning community to invoke GPU and achieved parallel computation for dynamics, making the simulator cross-platform and easy to develop. Two examples were presented to illustrate the functionality of the platform.

In the future, we plan to open-source the quadrotor part of the simulator, package it as a ROS node, and display it using RVIZ. We hope the proposed simulating architecture can provide valuable references for the design of next-generation large-scale swarm robotic simulators.


\bibliographystyle{ref/IEEEtran}
\bibliography{ref/IEEEabrv, ref/references}


\end{document}